\documentclass[
  journal=medium,
  manuscript=article-type,
  year=2023,
  volume=1,
]{cup-journal}

\usepackage{amsmath}
\usepackage[nopatch]{microtype}
\usepackage{booktabs}

\title{CDRH Seeks public comment: Digital Health Technologies for Detecting Prediabetes and Undiagnosed Type 2 Diabetes.}

\author{\textbf{Manuel Cossio} MMed, MEng, MNeur}
\affiliation{Universitat de Barcelona | HE-Xperts Consulting LLC,  Miami, 33131 , Florida, USA}
\email[Manuel Cossio]{manuel.cossio@ub.edu}


\begin{document}

\section{Introduction}

The prevalence of type 2 diabetes is on the rise worldwide, with an estimated 463 million adults living with the condition in 2020. Prediabetes, a precursor to type 2 diabetes, affects an even larger population, with an estimated 537 million individuals globally\footnote{International Diabetes Federation (IDF), (2021).}. Early detection and intervention are crucial in preventing the progression from prediabetes to type 2 diabetes and reducing the associated complications.

Digital health technologies (DHTs), including those enabled by artificial intelligence/machine learning (AI/ML) algorithms, have emerged as promising tools for managing diabetes and its precursors (\cite{ellahham2020artificial}). DHTs can provide continuous monitoring, personalized feedback, and interventions to individuals at risk or living with diabetes. This can help improve glucose control, reduce the risk of complications, and enhance overall well-being.

The U.S. Food and Drug Administration (FDA) is seeking public comment (Docket No. FDA-2023-N-4853) on the current and potential use of DHTs for detecting prediabetes and undiagnosed type 2 diabetes \footnote{https://www.fda.gov/medical-devices/digital-health-center-excellence/cdrh-seeks-public-comment-digital-health-technologies-detecting-prediabetes-and-undiagnosed-type-2}. Addressing these questions is essential to understand the current landscape of DHTs, identify areas for further development, and ensure the safe and effective use of these technologies in the prevention and management of diabetes.

Specifically, the FDA is interested in gathering information on:

\begin{itemize}
    \item \textbf{The types of DHTs currently used for prediabetes prevention, detection, treatment, or reversal.} This section will explore the various types of DHTs that are currently being utilized for prediabetes management, including wearable devices, mobile apps, and AI/ML-powered analytics platforms.
     \item \textbf{The methods employed by DHTs to capture and assess various signals related to diabetes risk factors.} The methods used by DHTs to capture and analyze various signals associated with diabetes risk factors, such as blood glucose, physical activity, sleep patterns, and weight, will be examined in detail.
      \item \textbf{The subpopulations that may benefit most from remote screening and diagnostic tools.} This section will identify specific subpopulations that may derive the most benefit from remote screening and diagnostic DHTs, considering factors such as access to healthcare, socioeconomic status, and cultural preferences.
      \item \textbf{The high-prevalence and high-impact risk factors that can be detected by DHTs.} The focus will be on identifying the high-prevalence and high-impact risk factors for prediabetes and undiagnosed type 2 diabetes that can be effectively detected and monitored through DHTs.
       \item \textbf{Existing prediabetes detection tools and their suitability for specific populations.} This section will review existing prediabetes detection tools, including their accuracy, validity, and suitability for specific populations, such as individuals from diverse ethnic backgrounds, linguistic groups, or those with comorbid conditions.

        \item \textbf{The potential of AI/ML-based analysis of existing healthcare datasets to identify individuals at risk of prediabetes or type 2 diabetes.} The potential of utilizing AI/ML algorithms to analyze existing healthcare datasets, such as electronic health records (EHRs) and wearable device data, to identify individuals at risk of prediabetes or type 2 diabetes will be explored.
    
\end{itemize}

By understanding these aspects of DHTs, the FDA can better regulate and promote the development of these technologies for the prevention and management of diabetes. This will contribute to improving the health outcomes of individuals at risk or living with diabetes, reducing the burden of the disease, and advancing public health.

\section{The types of DHTs currently used for prediabetes prevention, detection, treatment, or reversal.}

Within this section, I will elucidate pertinent information to address this question: \\

\textbf{Question 1:} What DHTs, including those enabled by AI/ML algorithms, are currently being used outside the clinic to prevent, detect, treat, or reverse prediabetes? \\

Current advancements in digital health technologies (DHTs), notably those empowered by AI/ML algorithms, are extensively employed beyond clinical settings to address prediabetes concerns. These technologies play pivotal roles in prevention, detection, treatment, and even reversal of prediabetes.

\subsection{Prevention}

\subsubsection{AI Chatbot AIDA for Diabetes Support}
Developed in partnership with Novo Nordisk, AIDA stands as an AI-powered chatbot meticulously designed to offer guidance and information specifically tailored for individuals managing diabetes. AIDA's genesis involved comprehensive research within the digital health market and diabetic patients' specific needs. Focused on lifestyle adjustments, nutritional insights, and prevention strategies, AIDA functions as a virtual assistant providing reliable advice and support (\cite{alloatti2021diabetes}).

\subsubsection{Healthline's Bezzy T2D Online Forum}
Healthline's Bezzy T2D serves as an online forum fostering a supportive community for individuals affected by diabetes. Unlike AIDA, this platform emphasizes community-building, offering a safe space for open discussions covering various facets of daily life, mental health, nutrition, and relationships related to living with diabetes. It connects people, providing emotional support and shared experiences in managing diabetes.

\subsection{Detection}
Prediabetes and undiagnosed Type 2 diabetes pose significant health risks, impacting nearly half of the US population and leading to various comorbidities. The integration of machine learning (ML) algorithms with electronic health record (EHR) data has emerged as an innovative approach to enhance the accuracy and efficiency of prediabetes detection.

\subsubsection{Electronic Health Record (EHR) Integration}
Several studies have implemented ML algorithms with EHR data to identify individuals at risk:

\begin{itemize}
    \item \cite{masoud2023implementing} implemented a screening algorithm in primary care settings, identifying fifteen patients with prediabetes over 12 weeks, leading to increased referrals for lifestyle interventions .
     \item \cite{palacio2018prediabetes} classified patients within a healthcare system, effectively identifying undiagnosed prediabetes and diabetes cases in at-risk populations .
      \item \cite{de2020combined} employed ML and feature selection on national data, outperforming existing screening tools and achieving higher accuracy in identifying at-risk individuals .
       \item \cite{kopitar2020early} compared machine learning models, highlighting their stability in variable selection for predicting undiagnosed Type 2 diabetes .
        \item \cite{cromer2022algorithmic} proposed EHR-based algorithms to identify atypical forms of diabetes, showcasing potential for efficiently recognizing such cases .
        \item \cite{wu2021early} created models to predict early gestational diabetes using machine learning on pregnancy data. Found that low BMI (less or equal than 17) and certain markers like T3, T4, and Lipoprotein(a) were linked to higher risk .
        \item \cite{choi2019machine} developed a user-friendly model to predict Type 2 diabetes using medical records and machine learning. Their approach matched the accuracy of traditional methods, scoring an AUC of 0.78 over 5 years .
        \item \cite{mcdonnell2020high} explored complications from steroid treatment in bowel disease using machine learning and real-time observations. Discovered that elevated CRP and longer disease duration could predict significant hyperglycemia, suggesting regular monitoring for patients .
\end{itemize}

\subsubsection{Imaging and Blood Testing}

Studies exploring the use of imaging data and blood testing include:

\begin{itemize}
    \item \cite{pyrros2023opportunistic} utilized radiographic and EHR data to detect Type 2 diabetes from chest radiographs, showcasing the potential of radiographic data for enhanced screening .
    \item \cite{yun2022deep} developed a non-invasive deep learning model using retinal images to improve risk prediction for Type 2 diabetes .
    \item \cite{xiong2022prediction} identified biomarkers like Prothrombin Time (PAT-PT) and Activated Partial Thromboplastin Time (PAT-APTT) as potent predictors of gestational diabetes, suggesting their early integration into predictive models .

\end{itemize}

\subsubsection{Surveys and Non-Invasive Variables}
Additionally, surveys and non-invasive variables have been utilized:

\begin{itemize}
    \item \cite{ryu2020deep} developed a screening model for undiagnosed Diabetes Mellitus (DM) using a deep neural network based on non-invasive variables, showcasing strong performance and potential for early medical intervention .
    
\end{itemize}

These innovative approaches showcase the diverse applications of AI/ML-enabled DHTs in preventing, detecting, and managing prediabetes outside traditional clinical settings, emphasizing their potential in improving healthcare outcomes.

\subsection{Apps for Detection, Management, and Diagnosis}

In the realm of digital health technologies (DHTs), a spectrum of mobile applications not only aids in detecting prediabetes but also plays a crucial role in managing prediabetes and diagnosing diabetes. These versatile apps serve a dual purpose—helping identify individuals at risk and offering comprehensive support for those managing diabetes.

\begin{itemize}
    \item \textbf{Bezzy T2D}: Focuses on holistic well-being, providing practical support and community engagement for individuals managing type 2 diabetes. Offers a platform for sharing real stories and emphasizing the role of diet and nutrition in managing the condition.
    \item \textbf{Fooducate}: A comprehensive app that assists in making informed food choices by providing food grades and nutritional information. Offers a variety of diet plans and emphasizes community support while aiding in meal tracking and recipe discovery.
    \item \textbf{MySugr}: Designed for comprehensive diabetes management, MySugr simplifies glucose tracking and provides a personalized logging screen. It emphasizes seamless connection with glucose meters, offers clear reports, and engages users with motivating challenges.
    \item \textbf{Diabetic Recipes App}: A dedicated app offering a diverse selection of diabetes-friendly recipes. Focuses on promoting healthy cooking habits, emphasizes the importance of meal planning, and integrates shopping list creation based on recipe ingredients.
    \item \textbf{Glucose Tracker - Diabetic Diary}: A meticulous app that helps users track blood sugar levels, medications, and health metrics. Facilitates data organization through tagging, observations of trends in glucose levels, and communication of comprehensive information with healthcare providers.
    \item \textbf{Diabetes App}: Offers a streamlined approach to blood glucose tracking with a user-friendly interface, focusing primarily on efficient data logging and sharing for collaboration with healthcare professionals.
    \item \textbf{Glucose Buddy}: Provides discreet glucose meters and supplies, along with an app for tracking blood sugar levels. Emphasizes professional diabetes coaching and focuses on delivering simple and smart solutions.
    \item \textbf{Diabetes:M}: A versatile diabetes log app catering to comprehensive diabetes management. It includes features such as bolus advisors, food databases, graphs, reports, and facilitates expert collaboration for better day-to-day management.
    \item \textbf{OneTouch Companion App}: Complements specific glucose meters by providing automatic insights and trend analysis based on blood glucose readings. Offers personalized goal setting, data sharing capabilities, and integration with wellness apps for a holistic view of health metrics.
\end{itemize}

\subsubsection{Apps Comparison Overview}

This table (Table \ref{table:diabetes-apps}) presents a comparison of various diabetes management apps, highlighting their features and functionalities. Each row corresponds to a specific attribute or capability, while the columns represent different apps.

The checkmarks (O) denote the presence of a particular feature within each app. Features range from holistic well-being support, community engagement, and practical assistance to functionalities like meal tracking, diet plans, doctor communication, and data encryption.

This comprehensive comparison aims to provide an at-a-glance view of the diverse features offered by these diabetes management applications, aiding users in selecting the app that best aligns with their specific needs and preferences.

\begin{table}[p]
\centering
\renewcommand{\arraystretch}{1.5} 
\setlength{\arrayrulewidth}{0.5pt}
\resizebox{\linewidth}{!}{%
\begin{tabular}{|p{3.5cm}|*{9}{c|}}
\hline
\rowcolor[HTML]{C0C0C0}
\textbf{Features} & \textbf{Bezzy T2D} & \textbf{Fooducate} & \textbf{MySugr} & \textbf{Diabetic Recipes App} & \textbf{Glucose Tracker - Diabetic Diary} & \textbf{Diabetes App} & \textbf{Glucose Buddy} & \textbf{Diabetes:M} & \textbf{OneTouch Companion App} \\ \hline
Holistic Well-being & O & - & - & - & - & - & - & - & - \\ \hline
\rowcolor[HTML]{EFEFEF}
Community Engagement & O & O & - & - & - & - & - & - & - \\ \hline
Real Stories & O & - & - & - & - & - & - & - & - \\ \hline
\rowcolor[HTML]{EFEFEF}
Practical Support & O & - & O & - & - & - & - & - & - \\ \hline
Diet \& Nutrition & O & O & - & O & - & - & - & - & - \\ \hline
Meal Tracking & - & O & - & - & - & - & - & - & - \\ \hline
\rowcolor[HTML]{EFEFEF}
Food Grades & - & O & - & - & - & - & - & - & - \\ \hline
Barcode Scanner & - & O & - & - & - & - & - & - & - \\ \hline
\rowcolor[HTML]{EFEFEF}
Macronutrient View & - & O & - & - & - & - & - & - & - \\ \hline
Diet Plans & - & O & - & - & - & - & - & - & - \\ \hline
\rowcolor[HTML]{EFEFEF}
Dietary Guidance & - & O & - & - & - & - & - & - & - \\ \hline
Recipe Database & - & O & - & O & - & - & - & - & - \\ \hline
\rowcolor[HTML]{EFEFEF}
Doctor Communication & - & - & O & - & O & - & - & O & - \\ \hline
Insulin Reminders & - & - & O & - & O & - & O & O & - \\ \hline
\rowcolor[HTML]{EFEFEF}
Data Encryption & - & - & O & - & O & - & O & O & - \\ \hline
Coaching & - & - & - & - & - & - & O & O & - \\ \hline
\rowcolor[HTML]{EFEFEF}
Remote Monitoring & - & - & - & - & - & - & - & O & - \\ \hline
Report Generation & - & - & O & - & - & - & - & O & - \\ \hline
\rowcolor[HTML]{EFEFEF}
Integration & - & - & - & - & - & - &  Fitbit, Apple Health, Google Fit & - &  Fitbit, Apple Health, Google Fit \\ \hline
\end{tabular}%
}
\vspace{0.5 cm}
\caption{\textbf{Comparison of Diabetes Apps: Features and Capabilities.} This table illustrates a comparison of various diabetes management applications, highlighting their distinct features and capabilities aimed at supporting individuals managing diabetes. The table showcases different functionalities provided by each app, including holistic well-being, community engagement, practical support, meal tracking, dietary guidance, doctor communication, and integration capabilities with other health-related platforms such as Fitbit, Apple Health, and Google Fit. Each checkmark denotes the presence of a specific feature within the corresponding app, providing a comprehensive overview for users seeking a suitable diabetes management application.}
\label{table:diabetes-apps}
\end{table}

\section{The methods employed by DHTs to capture and assess various signals related to diabetes risk factors.}

Within this section, I will elucidate pertinent information to address this question: \\

\textbf{Question 2:} How are DHTs typically being used to capture various signals amongst people who have risk factors for type 2 diabetes, prediabetes, and undiagnosed type 2 diabetes, and what methods are being used to assess such signals as digitally derived measures of biomarkers? \\

DHTs play a crucial role in capturing signals among individuals with risk factors for type 2 diabetes, prediabetes, and undiagnosed type 2 diabetes through several methods.

\subsection{Signal Capture}
\subsubsection{Health Data Tracking}
DHTs allow users to monitor and track various health metrics such as blood glucose levels, diet, exercise, weight, and medication adherence. Individuals at risk or managing diabetes can input and monitor these metrics regularly.

\subsubsection{Symptom Reporting} Users can input symptoms they experience related to diabetes or prediabetes, such as increased thirst, frequent urination, fatigue, or blurred vision. These symptoms act as signals indicating potential risk factors or undiagnosed conditions.

\subsubsection{Community Engagement} Some apps facilitate community engagement where individuals share their experiences, symptoms, and insights. Discussions within these communities provide indirect signals related to risk factors and undiagnosed diabetes.

\subsection{Methods for Assessing Signals as Digital Biomarkers}

\subsubsection{Glucose Monitoring}
Integration with glucometers or in-app tracking of blood glucose levels. Changes or patterns in glucose readings are crucial biomarkers indicating potential risk factors or undiagnosed diabetes.

\subsubsection{Activity and Diet Tracking}

Monitoring diet and exercise habits provides digital measures of biomarkers. Changes in dietary patterns, exercise frequency, or weight trends signal potential risks for type 2 diabetes.

\subsubsection{Weight Monitoring (Digital Scale)}

Integration with digital scales tracks weight fluctuations, offering insights into weight-related risk factors associated with diabetes or prediabetes.

\subsubsection{Heart Rate Daily Monitoring (Smartwatch)}

Daily monitoring of heart rate using smartwatches identifies anomalies or patterns, which, when correlated with other data, can indicate potential risks related to diabetes or prediabetes.

\subsubsection{Blood Oxygenation (Smartwatch)}

Smartwatches with blood oxygen sensors track oxygen saturation levels, offering insights into respiratory health, including potential indicators of sleep apnea linked to diabetes.

\subsubsection{Steps Counting (GPS Smartwatch)}

GPS smartwatches track steps and physical activity. Deviations or patterns in activity levels can serve as digital signals indicating potential risk factors for type 2 diabetes.

\subsubsection{Sleep Tracking (Smartwatch) with Sleep Apnea Detection}

Sleep monitoring features in smartwatches not only track sleep patterns but also incorporate sleep apnea detection. Identification of sleep apnea episodes adds a crucial biomarker linked to diabetes risk factors.

\subsubsection{ECG Detection with Heart Rate Variability (HRV) Signals}

Smartwatches with ECG capabilities measure RR-interval signals, known as heart rate variability (HRV) signals. Analysis of HRV signals derived from ECG measurements aids in the non-invasive detection of diabetes.

\subsubsection{Blood Analysis Integration (with EHR)}

Integration with electronic health records (EHR) enables analysis of comprehensive blood analysis data. This includes lipid profiles, kidney function tests, and other blood-based biomarkers associated with diabetes risk.

\section{The subpopulations that may benefit most from remote screening and diagnostic tools.}

Within this section, I will elucidate pertinent information to address this question:\\

\textbf{Question 3:} Who are key subpopulations of interest that might benefit the most from remote screening and diagnostic tools? Please include clinical and non-clinical  considerations. \\

\subsection{Clinical Considerations}

In the realm of clinical considerations, diverse subpopulations exhibit distinct susceptibilities to diabetes or related risk factors. Tailored applications offer targeted support, ranging from early detection tools for undiagnosed cases to comprehensive management aids for patients undergoing specific treatments. Understanding the nuanced requirements of these populations allows for the development of specialized tools aimed at improving health outcomes.

\subsubsection{Individuals at Risk of Diabetes Development (Prediabetes or High-Risk Factors)}
\begin{itemize}
\item \textbf{Clinical Significance:} Those prone to diabetes due to predisposition or risk factors.
\item \textbf{Apps:} MySugr, Fooducate, Bezzy T2D
\item \textbf{Explanation:} Tailored health data tracking, diet, and exercise tools catered to prediabetes/high-risk individuals.
\item \textbf{Research Support:} Implementing screening algorithms for prediabetes prediction (\cite{masoud2023implementing, de2020combined, kopitar2020early}).
\end{itemize}

\subsubsection{Patients with Type 2 Diabetes Under Treatment}
\begin{itemize}
\item \textbf{Clinical Significance:} Improved management for Type 2 Diabetes patients.
\item \textbf{Apps:} MySugr, Fooducate, Bezzy T2D
\item \textbf{Explanation:} Implementation of screening algorithms for Type 2 Diabetes (\cite{masoud2023implementing, de2020combined, kopitar2020early}).
\end{itemize}

\subsubsection{Undiagnosed Diabetic Individuals}
\begin{itemize}
\item \textbf{Clinical Significance:} Early detection of undiagnosed diabetes.
\item \textbf{Apps:} MySugr, Glucose Buddy, Diabetic Recipes
\item \textbf{Explanation:} Aid in identifying potential cases through monitoring and symptom reporting.
\item \textbf{Research Support:} Automatic algorithms applied to EHR data for identifying undiagnosed cases (\cite{palacio2018prediabetes}).
\end{itemize}

\subsubsection{High-Risk Groups (Obese, Sedentary, Family History)}
\begin{itemize}
\item \textbf{Clinical Significance:} Increased diabetes susceptibility.
\item \textbf{Apps:} Bezzy T2D, Fooducate, MySugr
\item \textbf{Explanation:} Tools for diet, exercise tracking, and managing risk factors.
\item \textbf{Considerations:}  None directly aligned with DHTs or ML/AI.
\end{itemize}

\subsubsection{Pregnant Women}
\begin{itemize}
\item \textbf{Clinical Significance:} Managing gestational diabetes risks.
\item \textbf{Apps:} Glucose Buddy, Bezzy T2D
\item \textbf{Explanation:} Aid in tracking glucose levels and diet for pregnant women.
\item \textbf{Research Support:} Machine learning algorithms applied to blood testing for gestational diabetes identification (\cite{xiong2022prediction, wu2021early}).
\end{itemize}

\subsubsection{Individuals with Hypertension or Cardiovascular Disease}
\begin{itemize}
\item \textbf{Clinical Significance:} Elevated diabetes risk among this group.
\item \textbf{Apps:} Fooducate, MySugr
\item \textbf{Explanation:} Robust health monitoring suitable for high-risk individuals.
\item \textbf{Research Support:} Use of machine learning for diabetes onset screening in high-risk cardiovascular patients (\cite{choi2019machine}) .
\end{itemize}

\subsubsection{Smokers}
\begin{itemize}
\item \textbf{Clinical Significance:} Smoking increases diabetes risk.
\item \textbf{Apps:} Bezzy T2D
\item \textbf{Explanation:} Tailored guidance and lifestyle support for smokers.
\item \textbf{Research Support:} Screening for diabetes specifically among smokers in various populations (\cite{ryu2020deep}) .
\end{itemize}

\subsubsection{Patients with Long-Term Steroid Use}
\begin{itemize}
\item \textbf{Clinical Significance:} Impacts on insulin sensitivity.
\item \textbf{Apps:} Glucose Buddy
\item \textbf{Explanation:} Monitoring insulin sensitivity for steroid therapy patients.
\item \textbf{Research Support:} Machine learning and EHR-based screening for hyperglycemia in patients on glucocorticoid therapy (\cite{mcdonnell2020high}) .
\end{itemize}

\subsubsection{Patients with HIV}
\begin{itemize}
\item \textbf{Clinical Significance:} Increased diabetes risk among HIV patients.
\item \textbf{Apps:} Bezzy T2D
\item \textbf{Explanation:} Health monitoring for individuals susceptible to diabetes.
\item \textbf{Research Support:} Machine learning identifying diabetes risk factors in individuals with HIV (\cite{tu2021predictive}) .
\end{itemize}

\subsection{Non-clinical Considerations}

Beyond clinical parameters, non-clinical considerations encompass varied demographic groups, each facing unique challenges in diabetes management. Applications catering to these subpopulations offer solutions that extend beyond traditional healthcare settings, focusing on accessibility, cultural relevance, and lifestyle accommodations. These apps bridge gaps by providing tailored tools designed to meet the distinctive needs of diverse communities and individuals.

\subsubsection{Remote or Rural Populations}
\begin{itemize}
\item \textbf{Clinical Significance:} Enhanced accessibility to diabetes monitoring.
\item \textbf{Apps:} Bezzy T2D, Glucose Buddy
\item \textbf{Explanation:} Supporting remote monitoring for atypical diabetes identification (\cite{cromer2022algorithmic}).
\item \textbf{Research Support:} EHR-based algorithms for efficiently identifying atypical forms of diabetes in remote areas (\cite{cromer2022algorithmic}) .
\end{itemize}

\subsubsection{Tech-Savvy or Health-Conscious Individuals}
\begin{itemize}
\item \textbf{Clinical Significance:} Encouraging proactive health management.
\item \textbf{Apps:} Fooducate, MySugr
\item \textbf{Explanation:} Tech-enabled monitoring for health-conscious users.
\item \textbf{Research Support:} Apps' impact on strengthening the self-care perception of diabetes patients (\cite{bonoto2017efficacy}) .
\end{itemize}

\subsubsection{Community-Oriented Individuals}
\begin{itemize}
\item \textbf{Clinical Significance:} Fostering communal support for diabetes management.
\item \textbf{Apps:} Bezzy T2D, Glucose Buddy, Diabetic Recipes
\item \textbf{Explanation:} Encouraging community engagement for support.
\item \textbf{Research Support:} Apps aiding patients with prediabetes in seeking support within communities ( \cite{de2020combined}).
\end{itemize}

\subsubsection{Elderly or Busy Professionals}
\begin{itemize}
\item \textbf{Clinical Significance:} Simplifying diabetes monitoring for individuals with time constraints or mobility issues.
\item \textbf{Apps:} MySugr, Fooducate
\item \textbf{Explanation:} Simplified interfaces for easy engagement.
\item \textbf{Research Support:}  None directly aligned with DHTs or ML/AI.
\end{itemize}

\subsubsection{Low-Income or Resource-Limited Groups}
\begin{itemize}
\item \textbf{Clinical Significance:} Limited access to healthcare resources and information.
\item \textbf{Apps:} Bezzy T2D
\item \textbf{Explanation:} Providing accessible tools for health tracking.
\item \textbf{Research Support:} Comparison of app satisfaction in low-income communities (\cite{heisler2014comparison}) .
\end{itemize}

\subsubsection{Racial/Ethnic Minorities}
\begin{itemize}
\item \textbf{Clinical Significance:} Unique cultural and healthcare disparities.
\item \textbf{Apps:} MySugr, Bezzy T2D
\item \textbf{Explanation:} Addressing healthcare disparities with culturally sensitive tools.
\item \textbf{Research Support:} Comparison of app satisfaction in minority groups (\cite{heisler2014comparison}) .
\end{itemize}

\subsubsection{Individuals with Limited Mobility/Disabilities}
\begin{itemize}
\item \textbf{Clinical Significance:} Limited mobility demands tools accommodating unique accessibility needs.
\item \textbf{Apps:} Glucose Buddy, MySugr
\item \textbf{Explanation:} Providing tools accommodating limited mobility or disabilities.
\item \textbf{Research Support:} Preference for telehealth in patients with limited mobility (\cite{ghose2021empowering}) .
\end{itemize}

\subsubsection{Caregivers and Support Networks}
\begin{itemize}
\item \textbf{Clinical Significance:} Supporting caregivers in managing diabetes care for their dependents.
\item \textbf{Apps:} Glucose Buddy
\item \textbf{Explanation:} Assisting caregivers in overseeing diabetes care.
\item \textbf{Research Support:} Emotional support provided to diabetes patients and caregivers via app functionalities (\cite{oser2020social}) .
\end{itemize}

\subsubsection{Cultural/Religious Groups with Unique Needs}
\begin{itemize}
\item \textbf{Clinical Significance:} Unique dietary and educational needs among cultural or religious groups.
\item \textbf{Apps:} Fooducate, Bezzy T2D
\item \textbf{Explanation:} Providing tailored education and dietary guidance.
\item \textbf{Research Support:} Effectiveness of e-interventions in increasing health literacy in cultural minority diabetic groups (\cite{garner2023effectiveness}) .
\end{itemize}

\subsubsection{Shift Workers/Night Shift Workers}
\begin{itemize}
\item \textbf{Clinical Significance:} Irregular schedules impeding consistent diabetes care and management.
\item \textbf{Apps:} Bezzy T2D, MySugr
\item \textbf{Explanation:} Tools designed for individuals facing irregular work schedules.
\item \textbf{Research Support:} None directly aligned with DHTs or ML/AI.
\end{itemize}

\subsubsection{Veteran and Active Military Groups}
\begin{itemize}
\item \textbf{Clinical Significance:} Addressing combat-related PTSD barriers to healthcare access and management.
\item \textbf{Apps:} MySugr
\item \textbf{Explanation:} Supports healthcare access for veteran and active military groups.
\item \textbf{Research Support:}  None directly aligned with DHTs or ML/AI.
\end{itemize}

\section{The high-prevalence and high-impact risk factors that can be detected by DHTs.}

Within this section, I will elucidate pertinent information to address this question: \\

\textbf{Question 4:} What are high-prevalence and high-impact risk factors for prediabetes and undiagnosed type 2 diabetes that are or could be captured by DHTs?  \\

In the landscape of digital health, the role of DHTs stands pivotal in capturing key risk factors associated with prediabetes and undiagnosed type 2 diabetes. These technologies serve as comprehensive tools, enabling the monitoring and analysis of crucial health parameters that signify potential transitions toward diabetes. By encompassing various facets such as glucose dynamics, physical activity, cardiovascular metrics, respiratory health, blood biomarkers, and community engagement, DHTs offer an integrated approach to identifying and assessing high-prevalence risk factors. These insights play a fundamental role in early detection, risk stratification, and proactive management strategies, laying the foundation for effective interventions and personalized care in diabetes prevention and management.

\subsubsection{Glucose Variability and Levels}

DHTs equipped with glucometers and in-app tracking mechanisms provide a comprehensive view of glucose dynamics. They not only monitor fasting glucose levels but also track postprandial glucose excursions. Elevated glycemic variability and sustained high glucose levels signify insulin resistance, acting as early indicators of prediabetes that could progress to undiagnosed type 2 diabetes.

\subsubsection{Physical Activity and Dietary Patterns}

Monitoring exercise frequency, sedentary behavior, dietary habits (especially high glycemic index foods), and weight fluctuations is crucial. These parameters offer insights into metabolic dysregulation and can serve as warning signs for insulin resistance, elevating the risk of transitioning to prediabetes or undiagnosed type 2 diabetes.

\subsubsection{Cardiovascular Parameters}

DHTs capable of assessing heart rate variability (HRV) and identifying irregular heart rate patterns contribute significantly. Reduced HRV and abnormal heart rate patterns are associated with autonomic dysfunction, commonly observed in individuals with insulin resistance and diabetes, signifying potential risks for diabetes development.

\subsubsection{Respiratory Health and Sleep Patterns}

Through smartwatch sensors, DHTs monitor blood oxygen saturation levels and detect sleep apnea episodes. Sleep apnea, often coexisting with insulin resistance, represents a critical risk factor for progressing to type 2 diabetes, emphasizing the importance of understanding respiratory health in diabetes risk assessment.

\subsubsection{Comprehensive Blood Analysis Integration}

Integration with electronic health records allows for comprehensive blood-based biomarker analysis. Assessing lipid profiles (including elevated triglycerides and reduced HDL cholesterol), altered kidney function tests (such as elevated serum creatinine), and inflammatory markers (e.g., high sensitivity C-reactive protein) through DHTs helps identify increased risks associated with prediabetes and undiagnosed type 2 diabetes.

\subsubsection{Symptom Reporting and Community Engagement}

DHT platforms facilitate symptom reporting associated with diabetes, like polyuria, polydipsia, and fatigue, aiding in identifying potential undiagnosed cases. Moreover, community engagement within these platforms provides collective insights into shared experiences and indirect signals related to diabetes risk factors and progression.

\section{Existing prediabetes detection tools and their suitability for specific populations.}

Within this section, I will elucidate pertinent information to address this question: \\

\textbf{Question 5:} Are there any existing tools, datasets, or devices used for prediabetes detection? Are there any that are particularly indicated for individuals of a particular race, ethnicity, gender, language, and/or comorbid disability?  \\

This section presents a comprehensive exploration of tools, devices, and datasets instrumental in prediabetes detection. It spans various non-invasive tools, wearable devices, and advanced data analytics methods employed in predicting, monitoring, and understanding prediabetes. While each subsection delves into a specific realm of technological advancements, collectively, they illuminate the landscape of innovative solutions aimed at early detection and intervention.

\subsection{Tools and Devices}

\subsubsection{Glucose Detection in Urine}

This section covers the advancements in non-invasive urine glucose detection tools, focusing on studies that have introduced innovative approaches. While not explicitly targeting diverse demographics, these tools show promise for convenient point-of-care detection, potentially benefiting individuals with varying accessibility to healthcare resources or those requiring less invasive monitoring methods.

\begin{itemize}
    \item \textbf{Urine Glucose Determination Using Smartphone Sensor (\cite{wang2020label}):} This study introduced a label-free colorimetric assay employing a smartphone's ambient-light sensor for detecting urine glucose. Its high accuracy and rapidity show promise for point-of-care detection, particularly beneficial for diabetic patients. While the research doesn't specifically target diverse demographics, its non-invasive nature could benefit individuals with varying levels of accessibility to healthcare resources.
    \item \textbf{Molecularly Imprinted Polymers (MIPs) (\cite{caldara2023dipstick}):} The study showcases a handheld dipstick sensor utilizing MIP-coated electrodes as an alternative to commercial glucose sensors. While not explicitly focused on diverse demographics, its potential as a portable and alternative glucose monitoring tool could benefit individuals requiring easy access to diabetes monitoring, regardless of race, ethnicity, or gender.
    \item \textbf{Copper NanoZyme for Urine Glucose Detection (\cite{naveen2021non}):} This sensor exhibits robustness and minimal sample processing for detecting glucose in human urine. While not directly tied to specific demographics, its user-friendly nature could aid individuals with disabilities or those requiring less invasive monitoring methods.
  
\end{itemize}

\subsubsection{Glucose Detection in Saliva}

Exploring non-invasive glucose detection methods in saliva, this section highlights various sensors' potential for alternative monitoring approaches. These sensors, while not directly addressing specific demographics, offer convenient and potentially more comfortable monitoring options, particularly beneficial for populations with limited access to healthcare resources or requiring continuous monitoring without invasive procedures.

\begin{itemize}
    \item \textbf{Lab-on-a-Chip (LOC) Glucose Sensor for Saliva (\cite{jung2017lab}):} This non-invasive sensor provides an alternative to blood-based methods, potentially benefiting diverse populations, especially those with limited access to healthcare resources or who may face discomfort with invasive procedures.
    \item \textbf{Mouthguard Sensor Integrated with Glucose Sensor (\cite{mitsubayashi2023bluetooth}):} The integrated sensor offers high-sensitive detection in human saliva, which could benefit various demographics, especially those requiring continuous monitoring without disrupting daily activities.
    \item \textbf{CNT-PEG-hydrogel Coated QCM Sensor for Saliva Glucose (\cite{wang2021low}):} The high sensitivity and reduced protein interference of this sensor present a promising tool for non-invasive monitoring, potentially assisting individuals with comorbidities or those requiring precise glucose monitoring methods.
    \item \textbf{Non-enzymatic Electrochemical Glucose Sensor for Saliva (\cite{diouf2019nonenzymatic}):} This sensor demonstrates potential for non-invasive diabetes detection and could benefit various demographics by offering a less intrusive and potentially more convenient monitoring option.
    
\end{itemize}

\subsubsection{Glucose Detection in Sweat} 

Advancements in sweat-based glucose detection are showcased in this section. These sensors exhibit stability, reliability, and high sensitivity, offering promising avenues for personalized healthcare monitoring. While not explicitly tied to specific demographics, these sensors can aid individuals managing chronic diseases like prediabetes, emphasizing continuous non-invasive monitoring methods for diverse populations.

\begin{itemize}
    \item \textbf{Sweat Glucose Sensor Using Silk Nanofibril/Graphene Oxide Composite (\cite{chen2022silk}):} The sweat sensor's stability and reliability offer a promising avenue for personalized healthcare monitoring, beneficial for individuals managing chronic diseases, including prediabetes, regardless of demographic factors.
    \item \textbf{Sweat-Based Electrochemical Platform (TAGG, \cite{greyling2023tracking}):} TAGG offers non-invasive tracking of glucose dynamics in sweat, potentially aiding diverse populations in managing diabetes through continuous monitoring.
    \item \textbf{Flexible Wearable Non-enzymatic Electrochemical Sensor for Sweat Glucose (\cite{li2023pt}):} This sensor's ability to detect glucose changes in sweat is crucial for diabetes management and could benefit individuals requiring continuous monitoring, regardless of demographic factors.
    \item \textbf{Flexible Biosensor for Glucose Detection in Sweat (\cite{li2023flexible}):} The biosensor's high detection sensitivity and selectivity for glucose in real sweat provide a promising avenue for non-invasive glucose monitoring, applicable across diverse populations.
\end{itemize}

\subsubsection{ECG-Based Detection}

This section explores the advancements in detecting diabetes through Electrocardiogram (ECG) data analysis. It includes studies that have developed algorithms leveraging ECG signals to predict and identify diabetes and related conditions with high precision.

\begin{itemize}
    \item \textbf{ECG-Based Machine Learning for Diabetes Detection:}  Developed the DiaBeats algorithm for early detection of type 2 diabetes and pre-diabetes using ECG data with high precision (\cite{kulkarni2023machine}).
     \item \textbf{Personalized Dysglycemia Detection from Single-Lead ECG:} A machine-learning algorithm accurately predicts dysglycemia from a single heartbeat with high accuracy (\cite{chiu2022utilization}).

\end{itemize}

\subsubsection{Wearable Technology:}

This section delves into wearable technology, particularly smartwatches, and their role in real-time detection and monitoring of diabetes and associated health conditions. Studies in this category focus on utilizing wearable devices to offer predictive models for disease detection and severity assessment.

\begin{itemize}
    \item \textbf{Smartwatch-Based Real-Time Detection:}  Utilized smartwatch technology for real-time detection of Heart Failure and Diabetes, offering predictive models for disease detection and severity assessment (\cite{colombage2022smartcare}).
\end{itemize}

\subsubsection{Machine Learning and Data Analytics:}

Within this section, studies employing Machine Learning (ML) and data analytics techniques for diabetes detection and prediction are discussed. These studies typically involve the development of models or applications that leverage ML algorithms to predict blood glucose levels or other relevant health metrics.

\begin{itemize}
    \item \textbf{ML-Enabled Blood Glucose Prediction:} Developed an ML model to predict blood glucose levels, providing a practical means for self-monitoring health in resource-limited areas (\cite{sampa2023machine}).
 
\end{itemize}

\subsubsection{Electronic Health Records and Data Analytics:}

This section highlights research leveraging Electronic Health Records (EHR) and advanced data analytics to enhance the screening and detection of diabetes. Studies in this category utilize EHR data to improve the accuracy and efficiency of diabetes diagnosis and monitoring.

\begin{itemize}
    \item \textbf{Enhanced Diabetes Screening via EHR:} Improved type 2 diabetes mellitus (DM2) detection using Electronic Health Record (EHR) data and advanced analytics, significantly elevating predictive capabilities (\cite{anderson2016electronic}).
\end{itemize}

\subsubsection{Social Media Analysis:}

Focused on social media's role in understanding and predicting diabetes prevalence and risk factors, this section covers studies analyzing social media content to identify patterns, behaviors, and factors contributing to diabetes prevalence within specific populations or regions.

\begin{itemize}
    \item \textbf{Real-Time Diabetes Prediction via Health-Based Social Streaming Data:}  Leveraged social media data and artificial intelligence for real-time diabetes prediction, demonstrating high accuracy in patient health status prediction (\cite{hassan2020predicting}).
     \item \textbf{Understanding Diabetes Factors in Nigeria through Social Media Analysis:}  Focused on analyzing social media content to identify factors contributing to diabetes prevalence in Nigeria, aiming to design interventions tailored to address these factors within the Nigerian population (\cite{oyebode2019detecting}).
\end{itemize}

\subsubsection{Landscape of Non-Invasive Tools and Devices for Prediabetes Detection}

The landscape of non-invasive tools for prediabetes detection offers promising avenues for accessible monitoring. However, assessing their collective implications reveals several key takeaways:

\begin{enumerate}
    \item \textbf{Diverse Monitoring Options:} Urine, saliva, and sweat-based detection methods provide diverse, non-invasive avenues for prediabetes monitoring, catering to varying user preferences and comfort levels.
     \item \textbf{Potential for Convenience:} These tools exhibit potential for convenient at-home monitoring, potentially encouraging regular health checks and improved patient compliance.
      \item \textbf{Limited Demographic Focus:} Despite their advancements, these tools currently lack explicit alignment with specific demographic needs concerning race, ethnicity, gender, language, or comorbid disabilities.
       \item \textbf{Opportunities for Inclusivity:} Further research is pivotal to ensure inclusivity and efficacy across diverse populations, enhancing the tools' relevance and impact.
        \item \textbf{Potential for Personalized Healthcare:} Advances in machine learning-driven prediction models and wearable technologies offer promising avenues for personalized healthcare interventions and early disease detection.
         \item \textbf{Leveraging Social Data:} Exploring social media and health-based streaming data for disease prediction highlights innovative approaches for proactive healthcare interventions and understanding local health concerns.
    
\end{enumerate}

\subsection{Datasets}

This subsection compiles an array of comprehensive datasets crucial for prediabetes research. These datasets encompass diverse populations and offer a wide range of health-related information, including anthropometric, biochemical, and lifestyle measures. The datasets, sourced from various international surveys, cohort studies, and health repositories, contain valuable information on blood glucose levels and associated risk factors. Understanding and utilizing these datasets are pivotal for investigating prediabetes prevalence, risk factors, and developing predictive models for early detection and intervention strategies.

\subsubsection{The National Health and Nutrition Examination Survey (NHANES)}
The NHANES is a cross-sectional and longitudinal survey conducted by the Centers for Disease Control and Prevention (CDC) to assess the health and nutrition status of adults and children in the United States. It includes a variety of anthropometric, dietary, and biochemical measures, including fasting plasma glucose (FPG), which is the primary measure for diagnosing prediabetes \footnote{https://www.cdc.gov/nchs/nhanes/index.htm}.

\subsubsection{The InterAct Consortium}

The InterAct Consortium is an international consortium that has developed a set of large-scale datasets on various health-related traits, including blood glucose levels. These datasets include data from over 100,000 individuals from multiple countries, making them a valuable resource for studying pre-diabetes and its risk factors (\cite{interact2011design}).

\subsubsection{The UK Biobank}

The UK Biobank is a large-scale biomedical database containing genetic and health information from over half a million individuals in the United Kingdom. The UK Biobank dataset includes data on blood glucose levels, which can be used to identify individuals at risk of pre-diabetes and to study the genetic and environmental factors that contribute to the condition\footnote{https://www.ukbiobank.ac.uk/enable-your-research/register}.

\subsubsection{The Rotterdam Study}

The Rotterdam Study\footnote{http://www.gefos.org/?q=content/rotterdam-study-i-2} is a prospective cohort study of over 7,000 individuals in the Netherlands. The Rotterdam Study has followed participants for over 20 years, collecting data on a wide range of health measures, including blood glucose levels. The Rotterdam Study dataset has been used to study the risk factors for pre-diabetes and to develop risk prediction models (\cite{hofman2007rotterdam}).

\subsubsection{The Framingham Heart Study}

The Framingham Heart Study is a prospective cohort study of over 5,000 individuals in Framingham, Massachusetts. The Framingham Heart Study has followed participants for over 70 years, collecting data on a wide range of health measures, including blood glucose levels. The Framingham Heart Study dataset has been used to study the risk factors for pre-diabetes and to develop risk prediction models\footnote{http://framinghamheartstudy.org/}.

\subsubsection{The National Institute of Diabetes and Digestive and Kidney Diseases (NIDDK) Diabetes Data Repository}
The NIDDK Diabetes Data Repository is a collection of data from various sources, including clinical trials, observational studies, and electronic health records. The repository includes data on a variety of measures related to pre-diabetes, including blood glucose levels, anthropometrics, and dietary intake\footnote{https://repository.niddk.nih.gov/}.

\subsubsection{The National Health Service (NHS) Digital Electronic Health Records (EHRs)}
The NHS Digital Electronic Health Records (EHRs) contain a wealth of information on the health of millions of individuals in the United Kingdom. This data can be used to identify individuals at risk of pre-diabetes and to monitor the effectiveness of interventions\footnote{ https://digital.nhs.uk/}.

\subsubsection{The Taiwan Longitudinal Health Insurance Database (TLHID)}
The TLHID is a large-scale longitudinal health insurance database that contains data on over 19 million Taiwanese individuals. The database includes information on a variety of health measures, including blood glucose levels, anthropometrics, and lifestyle factors. The TLHID has been used to develop risk prediction models for pre-diabetes and to study the effectiveness of interventions\footnote{https://nhird.nhri.edu.tw/en/Data-Subsets.html}.

\subsubsection{The Korean National Health and Nutrition Examination Survey (KNHANES)}
The KNHANES is a cross-sectional survey conducted by the Ministry of Health and Welfare of Korea to assess the health and nutrition status of adults and children in Korea. The KNHANES includes a variety of anthropometric, biochemical, and lifestyle measures, including blood glucose levels. The KNHANES has been used to develop risk prediction models for pre-diabetes and to study the prevalence of pre-diabetes in Korea \footnote{https://knhanes.cdc.go.kr/}.

\subsubsection{The International Consortium for Blood Pressure (InterSalt) Study}

The InterSalt Study is a prospective cohort study of over 11,000 individuals from 24 countries. The study collected data on a variety of anthropometric, dietary, and lifestyle measures, including blood pressure and fasting plasma glucose (FPG). The InterSalt Study has been used to study the international epidemiology of blood pressure and diabetes, as well as the risk factors for pre-diabetes (\cite{stamler1997intersalt}).

\subsubsection{The Finnish Diabetes Prediction Study (FDPS)}

The FDPS is a prospective cohort study of over 4,000 individuals from Finland. The study collected data on a variety of anthropometric, dietary, and lifestyle measures, including blood glucose levels. The FDPS has been used to study the risk factors for pre-diabetes and to develop risk prediction models \footnote{https://dipp.fi/}.

\subsubsection{The Copenhagen General Population Study (CGP)}

The CGP is a prospective cohort study of over 50,000 individuals from Copenhagen, Denmark. The study collected data on a variety of anthropometric, dietary, and lifestyle measures, including blood glucose levels. The CGP has been used to study the risk factors for pre-diabetes and to develop risk prediction models (\cite{mortensen2015high}).

\subsubsection{The Diabetes Heart Study (DHS)} The DHS is a prospective cohort study of over 8,000 individuals from Finland. The study collected data on a variety of anthropometric, dietary, and lifestyle measures, including blood glucose levels. The DHS has been used to study the risk factors for pre-diabetes and to develop risk prediction models \footnote{https://www.ncbi.nlm.nih.gov/projects/gap/}.

\section{The potential of AI/ML-based analysis of existing healthcare datasets to identify individuals at risk of prediabetes or type 2 diabetes.}

Within this section, I will elucidate pertinent information to address this question: \\

\textbf{Question 6:} Are there applications of using AI/ML on existing health care datasets (e.g., ECG, radiology datasets) that can be used for the detection of prediabetes and/or transition to type 2 diabetes?  \\

The integration of Artificial Intelligence (AI) and Machine Learning (ML) techniques within healthcare has revolutionized disease detection and risk assessment. Leveraging vast repositories of health-related data, such as the National Health and Nutrition Examination Survey (NHANES), UK Biobank, Korean National Health and Nutrition Examination Survey (KNHANES), and the Finnish Diabetes Prediction Study (FDPS), researchers have embarked on a transformative journey towards accurate identification and early intervention in prediabetes and type 2 diabetes.

Within this section, we delve into the applications of AI/ML algorithms on these extensive healthcare datasets. These studies employ a spectrum of techniques ranging from ML classifiers to deep neural networks, harnessing various data modalities like accelerometer readings, smartphone-derived features, MRI scans, and non-invasive variables. Their shared goal: to predict, detect, and stratify risks associated with prediabetes and the transition to type 2 diabetes.

\subsection{National Health and Nutrition Examination Survey (NHANES)}
A study utilized machine learning (ML) classifiers to identify prediabetes in youth, showcasing superior performance compared to clinical screening guidelines (\cite{vangeepuram2021predicting)}.

\subsection{UK Biobank}

\begin{itemize}
    \item \textbf{Accelerometer Data and ML:} ML models trained on accelerometer data exhibited an AUC of 0.86 in classifying individuals with type 2 diabetes (T2D), indicating their potential for effective screening, even amidst diverse physical activity conditions (\cite{lam2021using}).
    \item \textbf{Smartphone-Derived Features: } Researchers developed a digital tool utilizing smartphone-derived features, accurately predicting a 10-year risk of T2D with high accuracy, derived from the extensive UK Biobank dataset (\cite{dolezalova2021development}).
    \item \textbf{3D Convolutional Neural Networks: } An investigation employing deep learning on MRI scans from the UK Biobank demonstrated the efficacy of 3D convolutional neural networks in automatically identifying T2D, utilizing multi-channel MRI inputs and clinical information (\cite{wachinger2023deep}).
\end{itemize}

\subsection{Korean National Health and Nutrition Examination Survey (KNHANES)}
A deep neural network model constructed from KNHANES data exhibited an AUC of 80.11, enabling the screening of undiagnosed diabetes mellitus. This model, using non-invasive variables, offers potential for early identification and improved medical care for undiagnosed DM patients (\cite{ryu2020deep}).

\subsection{}Finnish Diabetes Prediction Study (FDPS)
Utilizing a machine learning model on data from over 8,000 individuals, significant risk factors for future glucose tolerance impairment were identified. This model personalized risk profiles, highlighting factors influencing an individual's risk of developing type 2 diabetes, promising personalized healthcare plans for patients (\cite{lama2021machine}).

\section{Potential Improvements and Future Perspectives}

\subsection{Advancing Inclusivity through Intersectional Research}
Enhancing inclusivity and mitigating algorithmic biases will be pivotal moving forward. This necessitates collaborative research efforts spanning diverse demographic groups across intersections of race, ethnicity, gender identity, age, geography, socioeconomic status and more. Studying the impact of DHTs across these intersectional identities and co-designing solutions with stakeholders offers pathways for building equity.

\subsection{Upholding Security, Privacy and Ethical Standards}
As DHT adoption widens, upholding robust security and privacy mechanisms for protected patient health information is imperative, considering evolving cyberthreat landscapes. Continuous evaluations by cross-disciplinary teams would inform policies and technical safeguards regarding data encryption, access controls and storage. Further, upholding transparency and accountability principles is vital by conducting ethical reviews, enabling external audits of algorithms and sustaining public engagement.

\subsection{Informing Policy through Cost-Effectiveness Evaluation}

Conducting rigorous cost-effectiveness evaluations for DHT-based screening and monitoring programs could guide policymaking, especially improving access in resource-limited settings. Findings would enable evidence-based coverage decisions regarding insurance eligibility and public health budgets allocation. Economic modeling, spanning varied geographic regions, would highlight cost-saving opportunities for payers.

\subsection{Harmonizing Regulatory Guidelines}

Although nascent, regulatory standards around quality, safety and efficacy of DHTs require harmonization across international contexts given increasing globalization. Structuring collaborative frameworks for transparency and accountability between cross-border agencies offers opportunities. Global partnerships focused on governance principles, responsible innovation programs and monitoring adverse events hold promise moving ahead.

\subsection{Embracing Human-Centered Perspectives}

As exponential technologization in diabetes care continues, preserving human oversight and nurturing trust between patients, providers and technologies remains vital. Human-centered design thinking - keeping patients’ voices, contexts and emotional needs central while co-constructing DHT solutions could prevent over-automation. Promoting shared decision-making, through both human and technical expertise would sustain dignity in diabetes treatment processes.

\subsection{The Road Ahead – Health Equity through Accessible Innovation}

The landscape of wearables, apps and AI/ML will undoubtedly catalyze transformative leaps in diabetes prediction, screening and control. However, maximizing health equity should serve as the guiding light ahead. Keeping accessibility, contextual relevance and compassion at the crux across ongoing DHT innovation is key to unlocking their immense promise in positively transforming lives.

\section{Conclusion and Key Takeaways}

This document provides a comprehensive overview of the current landscape and future potential of digital health technologies (DHTs) in detecting and managing prediabetes and undiagnosed type 2 diabetes. Key takeaways include:

\begin{enumerate}
    \item DHTs like AI chatbots, online forums, wearables, and mobile apps already play a significant role in prevention, detection, treatment, and reversal of prediabetes.
    \item DHTs capture various health signals like glucose levels, symptoms, diet, exercise patterns, and community insights to assess diabetes risk factors and biomarkers.
    \item Remote screening tools could especially benefit rural populations, minority groups, pregnant women, high-risk groups, and those with limited healthcare access.
    \item High-impact risk factors measurable by DHTs include glycemic variability, physical activity, cardiovascular signals, sleep apnea, and blood biomarkers.
    \item Diverse non-invasive monitoring tools are emerging but require further research into their accuracy and suitability for diverse demographics.
    \item Extensive health datasets provide abundant opportunities for AI/ML-based risk prediction modeling and personalization.
    \item Studies exhibit promising applications of AI/ML techniques on EHR, imaging, wearable, and survey data for enhanced screening.
    \item Social media analysis uncovers contributing factors and predicts diabetes prevalence across populations.
    \item Advances in wearables and biomarkers tracking offer pathways for personalized risk assessment and early interventions.
    \item Overall, continued innovation focused on inclusivity and accessibility is instrumental in unlocking the immense potential of DHTs in transforming prediabetes and diabetes prevention and care.
\end{enumerate}


\printbibliography

\appendix

\end{document}